\documentclass{article}
\usepackage{spconf,amsmath,epsfig}
\usepackage{amssymb}
\usepackage[breaklinks,colorlinks]{hyperref}
\usepackage{color}
\usepackage{xcolor}
\usepackage{pifont}
\usepackage{xspace}
\usepackage{booktabs}
\usepackage{graphicx}
\usepackage[normalem]{ulem}
\useunder{\uline}{\ul}{}

\newcommand{\lone}{{\large \ding{192}\xspace}}
\newcommand{\ltwo}{{\large \ding{193}\xspace}}
\newcommand{\port}{\!\,\texttt{Port\xspace}}
\newcommand{\greybox}{{\includegraphics[scale=0.04]{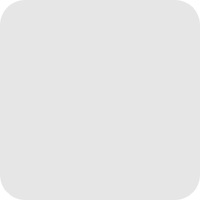}}}
\newcommand{\yellowbox}{{\includegraphics[scale=0.04]{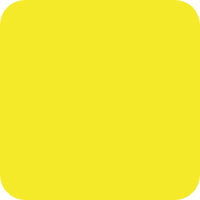}}}
\newcommand{\orangebox}{{\includegraphics[scale=0.04]{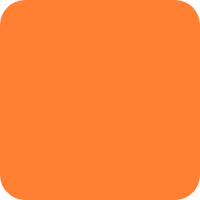}}}

\let\OLDthebibliography\thebibliography
\renewcommand\thebibliography[1]{
  \OLDthebibliography{#1}
  \setlength{\parskip}{0pt}
  \setlength{\itemsep}{0pt plus 0.3ex}
}

\begin{document}\sloppy

\def\x{{\mathbf x}}
\def\L{{\cal L}}

\title{Prompt When the Animal is: Temporal Animal Behavior Grounding with Positional Recovery Training}
%
\name{Sheng Yan$^{1}$, Xin Du$^{1}$, Zongying Li$^{1}$, Yi Wang$^{1}$, Hongcang Jin$^{1}$, Mengyuan Liu$^{2,\dagger}$\thanks{$\dagger$ Corresponding author: Mengyuan Liu (liumengyuan@pku.edu.cn).}}

\address{
$^1$School of Artificial Intelligence, Chongqing University of Technology\\
$^2$National Key Laboratory of General Artificial Intelligence, Shenzhen Graduate School, Peking University}

\maketitle

\begin{abstract}
Temporal grounding is crucial in multimodal learning, but it poses challenges when applied to animal behavior data due to the sparsity and uniform distribution of moments. To address these challenges, we propose a novel Positional Recovery Training framework (\port{}), which prompts the model with the start and end times of specific animal behaviors during training. Specifically, \port{} enhances the baseline model with a Recovering branch to reconstruct corrupted label sequences and align distributions via a Dual-alignment method. This allows the model to focus on specific temporal regions prompted by ground-truth information. Extensive experiments on the Animal Kingdom dataset demonstrate the effectiveness of \port{}, achieving an IoU@0.3 of 38.52. It emerges as one of the top performers in the sub-track of MMVRAC in ICME 2024 Grand Challenges.
\end{abstract}
\begin{keywords}
Temporal Grounding, Animal Kingdom, Multi-modal Learning, Deep Learning
\end{keywords}
\section{Introduction}
\label{sec:intro}

Temporal grounding, a crucial task in multimodal learning \cite{zhang2020span, gao2017tall, guo2022hybird, wang2023chan, wei2023conditional, lv2023temporal, li2024asmnet, zhao2024denoising, yan2023cross, ong2023chaotic, xu2023experts}, involves retrieving or localizing target moments that semantically correspond to given language queries. Recent literature \cite{ng2022animal} indicates that despite the commendable performance of representative works such as VSLNet \cite{zhang2020span} and LGI \cite{mun2020local} on conventional temporal grounding benchmarks \cite{krishna2017dense, gao2017tall}, they fall short when applied to animal behavior data \cite{ng2022animal}. We attribute this primarily to discrepancies in the sparsity of moments and their temporal position distribution:

\lone{} In the wild, capturing valuable footage of animals requires enduring long periods of waiting, where each captivating moment may vanish in an instant. This results in recorded videos containing only a fraction of animal footage or possibly none at all. In fact, as shown in Table \ref{tab:table1}, we list the Animal Kingdom dataset for animal behavior localization alongside the average moment, average duration, and the normalized moment length $\bar{L}_{m/v}$ relative to the length of the source video. As can be seen, Animal Kingdom's normalized moment length is only 0.19, which is a smaller characteristic when compared to traditional localization benchmarks \cite{krishna2017dense, gao2017tall}. Because of temporal sparsity, the shorter normalized length suggests difficulties with moment retrieval \cite{zhang2021natural}.

\begin{table}[t]
\small
\setlength{\tabcolsep}{2.2pt}
\centering
\begin{tabular}{@{}l|ccc@{}}
\toprule
Dataset & $\bar{L}_{video}$ & $\bar{L}_{moment}$ & $\bar{L}_{m/v}$ \\ \midrule
Charades-STA \cite{gao2017tall} &  30.59s & 8.22s & 0.27 \\
ActivityNet Captions \cite{krishna2017dense} & 117.61s & 36.18s & 0.32 \\
Animal Kingdom \cite{ng2022animal} & 38.15s & 6.27s & \textcolor{red}{0.19} \\ \bottomrule
\end{tabular}%
\vspace{-5pt}
\caption{Statistics of temporal grounding datasets, where $\bar{L}_{video}$, $\bar{L}_{moment}$ denotes the average length of videos, temporal moments in seconds, and $\bar{L}_{m/v}$ is the normalized moment length, against video length.}
\label{tab:table1}
\end{table}

\begin{figure}[t]
\centering
\vspace{-5pt}
\includegraphics[width=1.0\columnwidth]{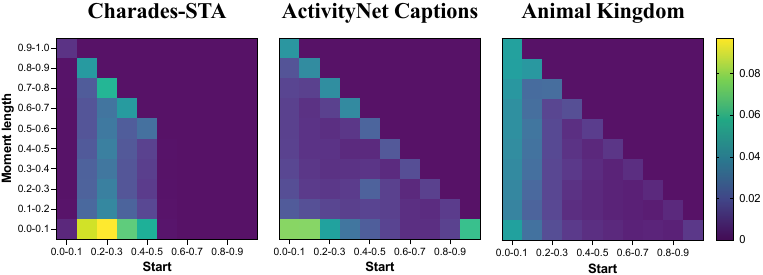}
\vspace{-25pt}
\caption{Distribution of temporal positions of target moments: Conventional grounding benchmarks VS. Animal Kingdom dataset. The color gradient indicates the relative frequency, with brighter shades indicating higher proportions.}
\label{fig:fig1}
\end{figure}

\ltwo{} The heatmaps in Fig. \ref{fig:fig1} illustrate the overall distribution of temporal moment positions between conventional temporal grounding benchmarks \cite{krishna2017dense, gao2017tall} and Animal Kingdom \cite{ng2022animal}. Here, the horizontal and vertical axes represent the starting time and duration of moments, respectively, normalized to the range of 0 to 1 by dividing them by the corresponding video lengths. Each grid cell denotes the corresponding percentage. In Charades-STA, target moments are more likely to start from the beginning of the video and last for approximately 10\% of the video length. For ActivityNet Captions, a peak can be observed near the bottom left corner of the graph. This peak corresponds to moments starting from the beginning of the video, also covering around 10\% of the video length. These biases provide strong priors for moment positions. Weighting candidate positions based on these distributions can increase the chances of obtaining the correct moments on conventional benchmarks \cite{otani2020uncovering}. However, such positional biases are significantly subtle in Animal Kingdom, as its distribution is more uniform (none of the grid cells exhibit bright green or yellow). Consequently, models benefiting from these positional biases \cite{zhang2020span, mun2020local, zhang2020learning} exhibit weakened capabilities.

To overcome the impact of the aforementioned moment disparities, {\ul we contemplate whether prompting the model with the starting or ending times of a certain animal behavior during training could focus the model's attention on that region.} To this end, we propose a Positional Recovery Training framework, inspired by recent works that injecting ground-truth information into object detection networks \cite{carion2020end, zhang2022dino, li2022dn}. Specifically, building upon the classical proposal-free framework, VSLNet \cite{zhang2020span}, we significantly enhance its final predictor. We divide the predictor into two branches. In addition to the \textit{Predicting branch}, which can predict the distribution of starting/ending boundaries of target moments normally, a \textit{Recovering branch} is introduced in parallel to perform positional recovery training. In this branch, our goal is to recover the flipped starting/ending label sequences. Since these sequences are already close to the ground truth, learning in this branch is easier and the predicted distributions are more accurate. Subsequently, a \textit{Dual-alignment} method is employed to force the distributions of the Predicting branch to overlap with them. This method prompts the Recovering branch to suggest the starting or ending times corresponding to a certain animal behavior to the Predicting branch.

Extensive experiments demonstrate that executing positional recovery training is effective. The proposed framework termed as \port{}, which stands for Temporal Animal Behavior Grounding with \textcolor{purple}{Po}sitional \textcolor{purple}{R}ecovery \textcolor{purple}{T}raining, achieves outstanding performance on the Animal Kingdom dataset with IoU@0.3=38.52. Moreover, it has been selected as one of the top performers in the Multi-Modal Video Reasoning and Analyzing Competition (MMVRAC) – Track 5: Video Grounding\footnote{\url{https://sutdcv.github.io/MMVRAC/}} at the International Conference on Multimedia and Expo in 2024.

\vspace{-5pt}
\section{Preliminary} \label{sec:plmy}
\vspace{-5pt}

In this paper, we adopt a subclass of proposal-free frameworks called Span-based prediction methods, which directly predict the probability of each video frame being the start/end position of a target moment. Next, we provide the definition of this task:

\noindent\textbf{Video and Text query}. A video is defined as a sequence of untrimmed frames, $\mathbf{V} \in {\mathbb{R}^{T \times d_{v}}}$, taken from a 3D ConvNet pre-trained on the Kinetics dataset \cite{carreira2017quo}, using RGB visual feature extraction and subsampling. For textual queries, they describe the target moment. This description contains precise animal behavior, such as \textit{``The bird dips its face into the water."} The data structure is a word sequence $\mathbf{Q} \in {{\mathbb{R}}^{N \times d_{w}}}$ of length $N$, where $d_{w}$ represents the word embedding dimension, indicating 300d GloVe vectors \cite{pennington2014glove}.

\noindent\textbf{Task objective}. Given a video $\mathbf{V}$ and a textual query $\mathbf{Q}$ describing the target moment, where $\tau_{s}$ and $\tau_{e}$ represent the ground-truth starting and ending time points, respectively. Span-based method aims to predict the starting and ending boundaries of the target moment span. Thus, it is necessary to map the starting/ending time $\tau_{s(e)}$ to the corresponding indices $i_{s(e)}$ in the video sequence $\mathbf{V}$. Specifically, assuming the video duration is $\mathfrak{T}$, the computation formula for the starting/ending indices $i_{s(e)}$ is: $i_{s(e)} = \left \langle \tau_{s(e)}/\mathfrak{T} \times T \right \rangle$, where $\left \langle \cdot \right \rangle$ denotes the rounding operator. Given a pair $(\mathbf{V},\mathbf{Q})$ as input, Span-based method locates the time span from the starting $i_{s}$ to the ending $i_{e}$. During inference, the predicted span boundaries can be easily converted to the corresponding times using $\tau_{s(e)}=i_{s(e)}/T \times \mathfrak{T}$.

\section{Method}

\begin{figure}[htbp]
\centering
\includegraphics[width=0.95\columnwidth]{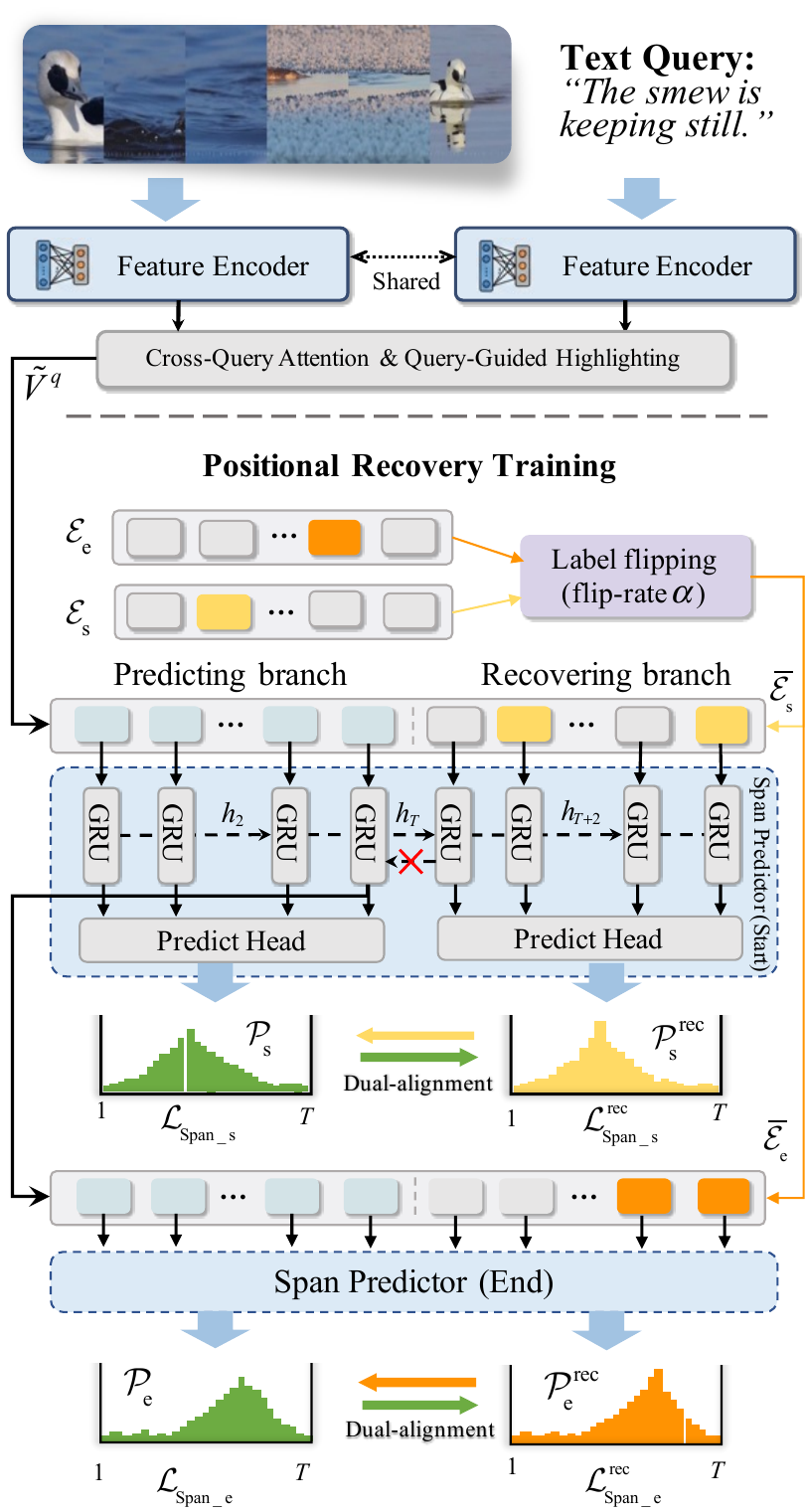}
\caption{Our proposed \port{} is built upon VSLNet. We significantly improve the predictor through Positional Recovery Training. We divide the predictor into two branches: the Predicting branch and the Recovering branch. Both branches share the same optimization objective, but the Recovering branch conducts recovery training on the flipped embedded sequences $\mathcal{\bar{E}}_{\text{s/e}}$ (composed of start(\yellowbox{})/non-start(\greybox{}) or end(\orangebox{})/non-end label embeddings).}
\label{fig:port}
\end{figure}

\subsection{VSLNet Baseline}
Our model is built upon VSLNet \cite{zhang2020span}, which treats the video as a text passage, the target moment as the answer span, and adapts a multimodal span-based question answering (QA) framework to the temporal grounding task. This architecture follows the ``encode-fuse-highlight-predict" paradigm (see Fig. \ref{fig:port} for an overview). Next, we briefly review the components of VSLNet and the training loss.

\noindent\textbf{Multi-modal learning}. The key components of this framework include the Feature Encoder, Context-Query Attention, and Conditional Span Predictor. The Feature Encoder utilizes a simplified version of the QANet \cite{yu2018fast} embedding encoder layer to project visual and textual features into a shared dimension. The Context-Query Attention captures cross-modal interactions between visual and textual features, which are crucial for understanding the video content within the context of the textual query. The Conditional Span Predictor is constructed based on a unidirectional LSTM and FFN layers to predict the starting and ending boundaries of answer spans in the video. Additionally, VSLNet introduces Query-Guided Highlighting (QGH) to enhance performance. QGH extends the boundaries of the target moment in the video, incorporating additional contextual information to aid in predicting the answer span. This binary classification module predicts the confidence of each visual feature belonging to the foreground or background.

To establish a solid baseline, in our framework, we adopt the subsequent work of VSLNet, self-guided parallel attention (SGPA) \cite{zhang2021parallel}, as our Feature Encoder. Moreover, we omit the extension of boundary in the QGH strategy. We note that these improvements are effective in enhancing the encoding effectiveness.

\noindent\textbf{VSLNet losses}.We keep the same base set of losses of \cite{zhang2020span}, defined as the weighted sum $\mathcal{L}_{\text{VSLNet}}=\mathcal{L}_{\text{span}}+\lambda_{\text{QGH}}\mathcal{L}_{\text{QGH}}$. In summary, the span prediction loss term $\mathcal{L}_{\text{span}}$ measures the difference between the predicted probability distributions of the starting and ending boundaries (via a cross-entropy loss). The QGH loss term $\mathcal{L}_{\text{QGH}}$ is related to the QGH strategy, which emphasizes highlighting relevant video features based on the input query. The loss is computed based on a binary classification task of predicting whether the visual features belong to the foreground (i.e, target moment) or background (with a binary cross-entropy loss). We set $\lambda_{\text{QGH}}$ to 5.0 in our experiments, as in \cite{zhang2020span}.

\subsection{Positional Recovery Training}

To address the moment disparities identified in Sec.~\ref{sec:intro}, we redesign the span predictor of the baseline with a two-branch architecture. The \textit{Predicting branch} performs standard boundary regression, while the \textit{Recovering branch} serves as a positional prompt: it receives slightly corrupted ground-truth label embeddings and is trained to reconstruct them, yielding well-localized distributions that guide the Predicting branch via \textit{Dual-alignment}. We describe the start boundary below; the end boundary is symmetric.

\noindent\textbf{Predicting branch.} The start boundary predictor comprises a unidirectional $\text{GRU}$ followed by a Predict Head—a 2-layer $\text{FFN}$ with $\text{LayerNorm}$ and $\text{ReLU}$. Given query-highlighted features $\mathbf{\widetilde{V}^{q}} \in \mathbb{R}^{T \times d}$ produced by the QGH strategy, the boundary scores are:
\begin{equation}
\mathcal{S}_{s} = \text{PredHead}(\mathbf{V_{s}^{q}}), \quad \mathbf{V_{s}^{q}} = \text{GRU}(\mathbf{\widetilde{V}^{q}})
\end{equation}
where $\mathcal{S}_{s} \in \mathbb{R}^{T}$. The probability distribution $\mathcal{P}_{s} = \text{Softmax}(\mathcal{S}_{s})$ is supervised by:
\begin{equation}
    \mathcal{L}_{\text{Span\_s}}= f_{\text{CE}}(\mathcal{P}_{s}, \mathcal{Y}_{s}) \label{eq:L_span}
\end{equation}
where $\mathcal{Y}_{s}$ is the start boundary label and $f_{\text{CE}}$ denotes cross-entropy.

\noindent\textbf{Recovering branch.} We embed the start labels into a sequence $\mathcal{E}_{s} \in \mathbb{R}^{T \times d}$ and apply \textit{label flipping}: a fraction $\alpha$ of start/non-start tokens are randomly exchanged, producing a corrupted sequence $\mathcal{\bar{E}}_{\text{s}}$. This corrupted sequence is concatenated with $\mathbf{\widetilde{V}^{q}}$ and fed into the same Span Predictor (Start), as shown in Fig.~\ref{fig:port}-middle. The $\text{GRU}$ accumulates context into its final hidden state $h_{T}$, which encodes sufficient sequence-level information to reconstruct the original label positions. The recovery objective is:
\begin{equation}
    \mathcal{L}_{\text{Span\_s}}^{\text{rec}}= f_{\text{CE}}(\mathcal{P}_{\text{s}}^{\text{rec}}, \mathcal{Y}_{s}) \label{eq:L_span_rec}
\end{equation}
where $\mathcal{P}_{\text{s}}^{\text{rec}} \in \mathbb{R}^{T}$ is the recovered start distribution. The two branches are executed in parallel, and the unidirectional $\text{GRU}$ prevents any ground-truth leakage into the Predicting branch.

\noindent\textbf{Dual-alignment.}
Since only a small fraction of labels are flipped, the Recovering branch produces sharper, more accurate distributions than the Predicting branch. We exploit this by minimizing the Kullback-Leibler divergence from the Predicting branch to the Recovering branch:
\begin{equation}
    \mathcal{L}_{\text{Align}}= D_{\text{KL}}(\mathcal{P}_{\text{s}}|| \mathcal{P}_{\text{s}}^{\text{rec}})
\end{equation}
The same procedure is applied to the end boundary, which is additionally conditioned on the predicted start (Fig.~\ref{fig:port}-bottom).

\noindent\textbf{Training loss.} The full objective of \port{} is $\mathcal{L}_{\text{VSLNet}}+\lambda_{\text{rec}}\mathcal{L}^{\text{rec}}_{\text{Span}}+\lambda_{\text{align}}\mathcal{L}_{\text{Align}}$, where $\mathcal{L}^{\text{rec}}_{\text{Span}}=\mathcal{L}^{\text{rec}}_{\text{Span\_s}}+\mathcal{L}^{\text{rec}}_{\text{Span\_e}}$, and $\lambda_{\text{rec}}$, $\lambda_{\text{align}}$ balance the two auxiliary terms.

\begin{table}[t]
\centering
\caption{Results on the Animal Kingdom test split.}
\label{tab:results}
\begin{tabular}{@{}lcccc@{}}
\toprule
Method & IoU=0.3 & IoU=0.5 & IoU=0.7 & mIoU \\ \midrule
LGI \cite{mun2020local} & 33.51 & 19.74 & 8.94 & 22.90 \\
VSLNet \cite{zhang2020span} &  33.74 & 20.83 & 12.22 & 25.02 \\
Port (Ours) & \textbf{38.52} & \textbf{26.41} & \textbf{15.87} &  \textbf{28.10} \\ \bottomrule
\end{tabular}%
\vspace{-5pt}
\end{table}

\section{Experiment}

\vspace{-5pt}
\subsection{Experiment Setting}

\noindent\textbf{Animal Kingdom} \cite{ng2022animal} provides frame-level natural language descriptions of wildlife behavior. It includes various wildlife species, such as whooper and hornbill, sourced from YouTube and ensured for quality through multiple rounds of inspection. The dataset comprises a total of 50 hours of video (across 4301 long video sequences) and 18,744 annotated sentences. Each video sequence contains 3-5 sentences. The dataset is divided into a training set (3489 video sequences and 14,995 text descriptions) and a test set (812 video sequences and 3749 text descriptions) in an 8:2 ratio.

\noindent\textbf{Evaluation protocol}. We adopt ``$\textrm{IoU}=\mu$" and ``mIoU" as evaluation metrics \cite{gao2017tall}. ``$\textrm{IoU}=\mu$" indicates the percentage of language queries with at least one retrieved moment having an Intersection over Union (IoU) greater than $\mu$ with the ground truth. ``mIoU" represents the average IoU across all test samples. In our experiments, we use $\mu \in \{0.3, 0.5, 0.7\}$ and retrieve the moments with the maximum joint probability of starting/ending boundaries (Eq. \ref{eq:L_span}) generated from boundary transformations mentioned in Sec. \ref{sec:plmy} (i.e., $\textrm{Recall}@1$).

\noindent\textbf{Implementation Details}. We use the AdamW optimizer to guide the optimization process with a batch size of 16 and a weight decay parameter of $0.01$. The model training process spans 100 epochs, with an initial learning rate of $2e-4$ linearly decayed. We set the maximum video length $T$ to 128 and the model's hidden dimension $d$ to 256. The flip rate $\alpha$ for positional recovery training is set to 0.2. The loss balancing parameters $\lambda_{\text{Recovering}}$ and $\lambda_{\text{Align}}$ are both set to 1.0. All experiments are conducted on a single NVIDIA RTX 4090.

%

\begin{table}[t]
\centering
\caption{Ablation study of Positional Recovery Training (PRT).}
\label{tab:results2}
\begin{tabular}{@{}lcccl@{}}
\toprule
Method  & IoU=0.5 & IoU=0.7 & mIoU \\ \midrule
Port  & 26.41 & 15.87 & 28.10 \\
\quad w/o Dual-alignment  & 25.37 & 15.52 & 26.91 \\
\quad w/o PRT  & 24.97 & 15.39 & 27.16 \\ \bottomrule
\end{tabular}
\vspace{-10pt}
\end{table}

\begin{table}[t]
\centering
\caption{Ablation study of Positional Encoding.}
\label{tab:pos}
\begin{tabular}{@{}lcccl@{}}
\toprule
Position encoding & IoU=0.5 & IoU=0.7 & mIoU \\ \midrule
Learned embeddings  & 24.72 & 13.77 & 26.54  \\
Sinusoidal encoding  & 25.42 & 14.49 & 26.25 \\
None & 26.41 & 15.87 & 28.10 \\ \bottomrule
\end{tabular}
\vspace{-5pt}
\end{table}


\begin{table}[t]
\centering
\caption{Ablation on various model hidden size ($d$)}
\label{tab:hiddensize}
\begin{tabular}{@{}lcccl@{}}
\toprule
Hidden dim & IoU=0.5 & IoU=0.7 & mIoU \\ \midrule
$d=128$  & 24.94 & 15.26 & 27.77 \\
$d=256$  & 26.41 & 15.87 & 28.10 \\
$d=384$ & 25.50 & 15.26 & 26.87 \\
$d=512$  & 25.45 & 15.82 & 26.96 \\ \bottomrule
\end{tabular}
\vspace{-5pt}
\end{table}

\vspace{-5pt}
\subsection{Quantitative Results} 
\vspace{-5pt}

We compare \port{} with existing methods documented in the literature, including LGI \cite{mun2020local} and VSLNet \cite{zhang2020span}. As shown in Table \ref{tab:results}, the best performance is highlighted in \textbf{bold}. Thanks to positional recovery training, i.e., by prompting the model with the start/end times of animal behaviors, our \port{} achieves the best performance.

\vspace{-5pt}
\subsection{Ablation Study} 
\vspace{-5pt}

\noindent\textbf{Ablation on Positional Recovery Training (PRT)}. Table \ref{tab:results2} presents the results without Dual-alignment (abbreviated as \textit{w/o Dual-alignment}) and without Positional Recovery Training (\textit{w/o PRT}). When \textit{w/o PRT}, our model degenerates into the VSLNet baseline. The ablation results indicate the effectiveness of our PRT (26.41 vs. 24.97 at IoU@0.5). An interesting observation is that the difference between the results of \textit{w/o Dual-alignment} and \textit{w/o PRT} is not significant. This suggests that the Dual-alignment method is crucial for PRT, indicating that PRT only works effectively through this method after prompting the model with start/end times.

\noindent\textbf{Is positional encoding all you need?} We examined the descriptions of animal behaviors and found them to be shorter than those of conventional localization benchmarks. Importantly, they rarely include words depicting temporal relationships, such as `before' or `after'. We speculate that strong temporal relationship modeling may be weakly correlated with Animal Kingdom. Therefore, we experimented with removing positional encodings and compared them with learnable and sine positional encodings. Table \ref{tab:pos} shows that no positional encoding is more suitable for temporal animal behavior grounding.

\noindent\textbf{Which hidden dim is the best?} The results in Table \ref{tab:hiddensize} demonstrate that the performance is optimal when the hidden dimension $d$ of the model is set to 256.

\begin{figure}[htbp]
\centering
\vspace{-5pt}
\includegraphics[width=1.0\columnwidth,trim=6 0 6 0,clip]{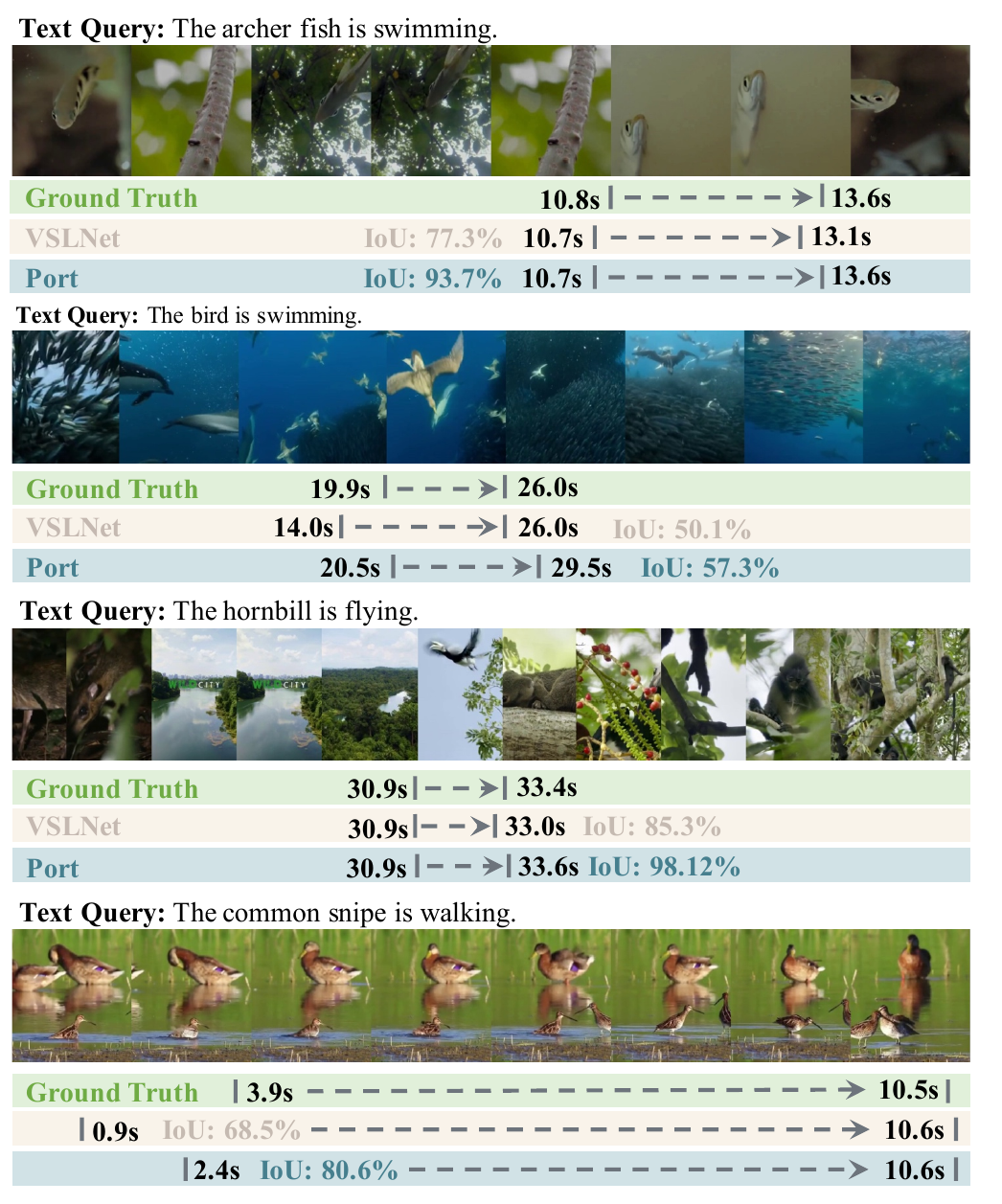}
\vspace{-25pt}
\caption{Visualization of the ground-truth moment and predictions by competitor.}
\label{fig:vesus}
\vspace{-20pt}
\end{figure}

\begin{figure}[t]
\centering
\vspace{-5pt}
\includegraphics[width=1.0\columnwidth,trim=2 0 2 0,clip]{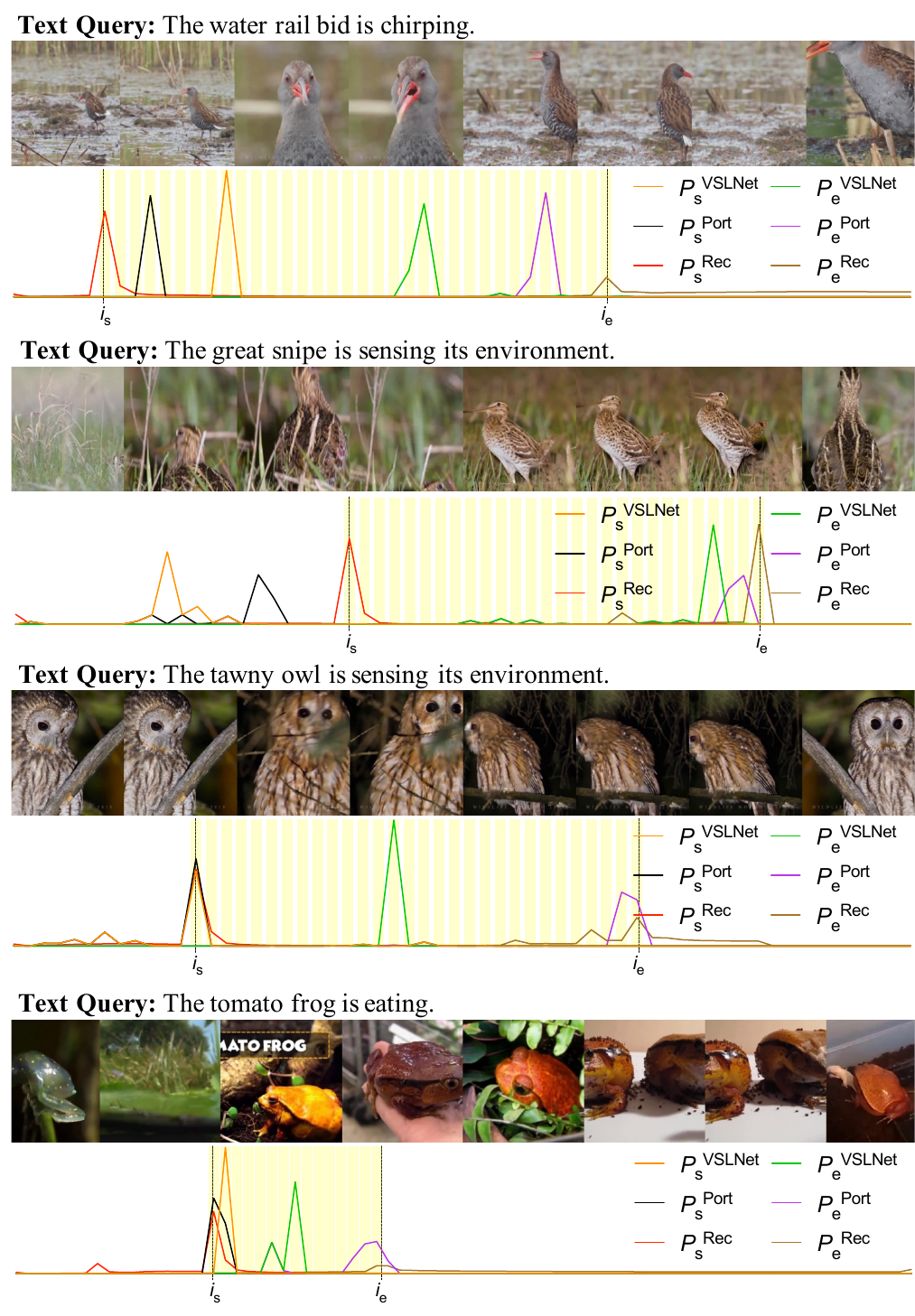}
\vspace{-25pt}
\caption{Visualization of the predicted distributions of the Predicting branch and the Recovering branch in \port{}, compared to the predicted distribution of VSLNet, with the moment regions highlighted in yellow.}
\label{fig:vesus2}
\vspace{-10pt}
\end{figure}

\subsection{Qualitative Results} 
\vspace{-5pt}

We visualize some qualitative results and compare \port{} with VSLNet. The results are shown in Fig. \ref{fig:vesus}. Regardless of short or long videos ($>$30s), our \port{} achieves higher IoU. However, our method also has limitations. In the second case, our model only achieves a 57\% IoU. For long videos, our method is currently unstable. This instability arises because in our experiments, long videos are compressed to a fixed length of 128. In this scenario, even small variations in the localized span indices can result in significant changes in the corresponding times after transformation.

Furthermore, we visualize the predicted distributions of the Predicting branch and the Recovering branch in \port{}. As shown in Fig. \ref{fig:vesus2}, the start and end probabilities of these two branches are represented as $P_{\text{s/e}}^{\text{Port}}$ and $P_{\text{s/e}}^{\text{Rec}}$, respectively, corresponding to the symbols $\mathcal{P}_{s/e}$ (Eq. \ref{eq:L_span}) and $\mathcal{P}_{s/e}^{rec}$ (Eq. \ref{eq:L_span_rec}) in the formulas. We compare them with the start and end probabilities of the VSLNet, $P_{\text{s/e}}^{\text{VSLNet}}$. The four cases in Fig. \ref{fig:vesus2} first demonstrate that the peaks of the Recovering branch distribution are very close to the start or end indices $i_{s/e}$ because we only perform slight label flipping, making learning in this branch easy and the predicted distribution accurate. Secondly, compared to the predicted distribution of VSLNet, the predicted distribution of the Predicting branch in \port{} is closer to the distribution of the Recovering branch. This demonstrates that Positional Recovery Training helps the model focus on the start/end regions, leading to more accurate grounding results.

\vspace{-5pt}
\section{Conclusion} 
\vspace{-5pt}

In this study, we investigated the emerging problem of temporal animal behavior grounding. We analyzed its discrepancies from conventional temporal grounding benchmarks, which are reflected in the sparsity and uniform distribution of temporal positions. To address this, we proposed the Positional Recovery Training framework (\port{}), which bridges this gap effectively by prompting the model with start and end times of certain animal behaviors.

In future work, we may consider leveraging LLM to identify the subject animals of the moments and adding a classification branch network to enhance model robustness.

\vspace{-5pt}
\section{Acknowledgement} 
\vspace{-5pt}

This work was supported by National Natural Science Foundation of China (No. 62203476), Natural Science Foundation of Shenzhen (No. JCYJ20230807120801002).


\small
\bibliographystyle{IEEEbib}
\bibliography{icme2023template}

@inproceedings{ng2022animal,
  title={Animal kingdom: A large and diverse dataset for animal behavior understanding},
  author={Ng, Xun Long and Ong, Kian Eng and Zheng, Qichen and Ni, Yun and Yeo, Si Yong and Liu, Jun},
  booktitle={Proceedings of the IEEE/CVF Conference on Computer Vision and Pattern Recognition},
  pages={19023--19034},
  year={2022}
}

@inproceedings{krishna2017dense,
  title={Dense-captioning events in videos},
  author={Krishna, Ranjay and Hata, Kenji and Ren, Frederic and Fei-Fei, Li and Carlos Niebles, Juan},
  booktitle={Proceedings of the IEEE international conference on computer vision},
  pages={706--715},
  year={2017}
}

@inproceedings{gao2017tall,
  title={Tall: Temporal activity localization via language query},
  author={Gao, Jiyang and Sun, Chen and Yang, Zhenheng and Nevatia, Ram},
  booktitle={Proceedings of the IEEE international conference on computer vision},
  pages={5267--5275},
  year={2017}
}

@article{zhang2020span,
  title={Span-based localizing network for natural language video localization},
  author={Zhang, Hao and Sun, Aixin and Jing, Wei and Zhou, Joey Tianyi},
  journal={arXiv preprint arXiv:2004.13931},
  year={2020}
}

@inproceedings{mun2020local,
  title={Local-global video-text interactions for temporal grounding},
  author={Mun, Jonghwan and Cho, Minsu and Han, Bohyung},
  booktitle={Proceedings of the IEEE/CVF Conference on Computer Vision and Pattern Recognition},
  pages={10810--10819},
  year={2020}
}

@article{otani2020uncovering,
  title={Uncovering hidden challenges in query-based video moment retrieval},
  author={Otani, Mayu and Nakashima, Yuta and Rahtu, Esa and Heikkil{\"a}, Janne},
  journal={arXiv preprint arXiv:2009.00325},
  year={2020}
}

@article{zhang2021natural,
  title={Natural language video localization: A revisit in span-based question answering framework},
  author={Zhang, Hao and Sun, Aixin and Jing, Wei and Zhen, Liangli and Zhou, Joey Tianyi and Goh, Rick Siow Mong},
  journal={IEEE transactions on pattern analysis and machine intelligence},
  volume={44},
  number={8},
  pages={4252--4266},
  year={2021},
  publisher={IEEE}
}

@inproceedings{guo2022hybird,
  title={A hybird alignment loss for temporal moment localization with natural language},
  author={Guo, Chao and Liu, Daizong and Zhou, Pan},
  booktitle={2022 IEEE International Conference on Multimedia and Expo (ICME)},
  pages={1--6},
  year={2022},
  organization={IEEE}
}

@inproceedings{zhang2020learning,
  title={Learning 2d temporal adjacent networks for moment localization with natural language},
  author={Zhang, Songyang and Peng, Houwen and Fu, Jianlong and Luo, Jiebo},
  booktitle={Proceedings of the AAAI Conference on Artificial Intelligence},
  volume={34},
  number={07},
  pages={12870--12877},
  year={2020}
}

@inproceedings{wang2023chan,
  title={CHAN: Cross-Modal Hybrid Attention Network for Temporal Language Grounding in Videos},
  author={Wang, Wen and Zhong, Ling and Gao, Guang and Wan, Minhong and Gu, Jason},
  booktitle={2023 IEEE International Conference on Multimedia and Expo (ICME)},
  pages={1499--1504},
  year={2023},
  organization={IEEE}
}

@inproceedings{wei2023conditional,
  title={Conditional Video-Text Reconstruction Network with Cauchy Mask for Weakly Supervised Temporal Sentence Grounding},
  author={Wei, Jueqi and Xu, Yuanwu and Chen, Mohan and Zhang, Yuejie and Feng, Rui and Gao, Shang},
  booktitle={2023 IEEE International Conference on Multimedia and Expo (ICME)},
  pages={1511--1516},
  year={2023},
  organization={IEEE}
}

@inproceedings{lv2023temporal,
  title={Temporal-enhanced Cross-modality Fusion Network for Video Sentence Grounding},
  author={Lv, Zezhong and Su, Bing},
  booktitle={2023 IEEE International Conference on Multimedia and Expo (ICME)},
  pages={1487--1492},
  year={2023},
  organization={IEEE}
}

@inproceedings{li2022dn,
  title={Dn-detr: Accelerate detr training by introducing query denoising},
  author={Li, Feng and Zhang, Hao and Liu, Shilong and Guo, Jian and Ni, Lionel M and Zhang, Lei},
  booktitle={Proceedings of the IEEE/CVF Conference on Computer Vision and Pattern Recognition},
  pages={13619--13627},
  year={2022}
}

@article{zhang2022dino,
  title={Dino: Detr with improved denoising anchor boxes for end-to-end object detection},
  author={Zhang, Hao and Li, Feng and Liu, Shilong and Zhang, Lei and Su, Hang and Zhu, Jun and Ni, Lionel M and Shum, Heung-Yeung},
  journal={arXiv preprint arXiv:2203.03605},
  year={2022}
}

@inproceedings{carreira2017quo,
  title={Quo vadis, action recognition? a new model and the kinetics dataset},
  author={Carreira, Joao and Zisserman, Andrew},
  booktitle={proceedings of the IEEE Conference on Computer Vision and Pattern Recognition},
  pages={6299--6308},
  year={2017}
}

@inproceedings{pennington2014glove,
  title={Glove: Global vectors for word representation},
  author={Pennington, Jeffrey and Socher, Richard and Manning, Christopher D},
  booktitle={Proceedings of the 2014 conference on empirical methods in natural language processing (EMNLP)},
  pages={1532--1543},
  year={2014}
}

@article{zhang2021parallel,
  title={Parallel attention network with sequence matching for video grounding},
  author={Zhang, Hao and Sun, Aixin and Jing, Wei and Zhen, Liangli and Zhou, Joey Tianyi and Goh, Rick Siow Mong},
  journal={arXiv preprint arXiv:2105.08481},
  year={2021}
}

@inproceedings{yu2018fast,
  title={Fast and accurate reading comprehension by combining self-attention and convolution},
  author={Yu, Adams Wei and Dohan, David and Le, Quoc and Luong, Thang and Zhao, Rui and Chen, Kai},
  booktitle={International conference on learning representations},
  volume={2},
  number={1},
  year={2018}
}

@article{li2024asmnet,
  title={ASMNet: Action and Style-Conditioned Motion Generative Network for 3D Human Motion Generation},
  author={Li, Zongying and Wang, Yong and Du, Xin and Wang, Can and Koch, Reinhard and Liu, Mengyuan},
  journal={Cyborg and Bionic Systems},
  volume={5},
  pages={0090},
  year={2024},
  publisher={AAAS}
}

@inproceedings{zhao2024denoising,
  title={Denoising Diffusion Probabilistic Models for Action-Conditioned 3D Motion Generation},
  author={Zhao, Mengyi and Liu, Mengyuan and Ren, Bin and Dai, Shuling and Sebe, Nicu},
  booktitle={ICASSP 2024-2024 IEEE International Conference on Acoustics, Speech and Signal Processing (ICASSP)},
  pages={4225--4229},
  year={2024},
  organization={IEEE}
}

@inproceedings{yan2023cross,
  title={Cross-Modal Retrieval for Motion and Text via DropTriple Loss},
  author={Yan, Sheng and Liu, Yang and Wang, Haoqiang and Du, Xin and Liu, Mengyuan and Liu, Hong},
  booktitle={Proceedings of the 5th ACM International Conference on Multimedia in Asia},
  pages={1--7},
  year={2023}
}

@inproceedings{ong2023chaotic,
  title={Chaotic World: A Large and Challenging Benchmark for Human Behavior Understanding in Chaotic Events},
  author={Ong, Kian Eng and Ng, Xun Long and Li, Yanchao and Ai, Wenjie and Zhao, Kuangyi and Yeo, Si Yong and Liu, Jun},
  booktitle={Proceedings of the IEEE/CVF International Conference on Computer Vision},
  pages={20213--20223},
  year={2023}
}

@article{xu2023experts,
  title={Experts Collaboration Learning for Continual Multi-Modal Reasoning},
  author={Xu, Li and Liu, Jun},
  journal={IEEE Transactions on Image Processing},
  year={2023},
  publisher={IEEE}
}

@inproceedings{carion2020end,
  title={End-to-end object detection with transformers},
  author={Carion, Nicolas and Massa, Francisco and Synnaeve, Gabriel and Usunier, Nicolas and Kirillov, Alexander and Zagoruyko, Sergey},
  booktitle={European conference on computer vision},
  pages={213--229},
  year={2020},
  organization={Springer}
}

\end{document}